# A study of influential factors in designing self-reconfigurable robots for green manufacturing

**Research-in-progress**

**Mahdi Fahmideh**
School of Information, Systems and Modelling
University of Technology Sydney
Australia, Sydney
Email: Mahdi.Fahmideh@uts.edu.au

**Thorsten Lammers**
School of Information, Systems and Modelling
Australia, Sydney
Email: Thorsten.Lammers@uts.edu.au

## Abstract

There is incremental growth in adopting self-reconfigurable robots in automating manufacturing conventional product lines. Using this class of robots adapting themselves with ever-changing environmental conditions has been acclaimed as a promising way of reducing energy consumption and environmental impact and thus enabling green manufacturing. Whilst the majority of existing research focuses on highlighting the efficacy of self-reconfigurable robots in energy reduction with technical driven solutions, the research on exploring the salient factors in design and development self-reconfigurable robots that directly enable or hinder green manufacturing is non-extant. This interdisciplinary research contributes to the nascent body of the knowledge by empirical investigation of design-time, run-time, and hardware aspects which should be contingently balanced when developing green-aware self-reconfigurable robots.

**Keywords** Green manufacturing, self-reconfigurable robots, robot design, green awareness

# 1 Introduction

The idea of *green manufacturing* is stimulated by increasing CO2 gas emissions, scarcity of raw material resources, and increasing levels of air pollution. Green manufacturing refers to a new manufacturing paradigm in which various green principles, policies, strategies, and technologies are employed to become more environmentally friendly and substitute input production materials with non-toxic and renewable ones Deif (2011). The notion of green manufacturing ranges from green purchasing to receiving customer orders and delivering the product Zhu and Sarkis (2004). Green manufacturing is augmented by green technologies along with more environmentally friendly product designs. By 2020, 60% of plant floor workers at top 2,000 ranked public companies in the world will work alongside automated assistance technologies such as robotics, 3D printing, artificial intelligence, and virtual reality Knickle (2016). Self-reconfigurable robotics is a new and emerging field of robotic systems. They consist of modular components that can automatically change their configuration to achieve different tasks based on environmental conditions.

The design and development of self-reconfigurable robots in alignment with the requirements of green manufacturing such as less energy consumption is a challenging task Brossog et al. (2015). In fact, adherence to green manufacturing is undergone by reducing the energy consumption of robots, which enable and support manufacturing processes. Hence, the design of green-aware self-reconfigurable robots can contribute to more energy-efficient production lines. An understanding of the relationship between key factors in the design and development of self-reconfigurable robotic systems that are important to address green requirements would significantly help to maximise the business impact of this technology. It would not only allow for a more efficient utilisation of robotic systems, but could also increase manufacturers' capabilities to design, develop, and increase the usability of self-reconfigurable robots for the service and consulting sectors, which play an important role to get a competitive advantage such as productivity, safety, and saving. To close this gap, the goal of this research is to answer the following research question: "What are key influential factors to be taken into account in a self-reconfigurable robot design with respect to green manufacturing requirements?" The key contributions of this study are to (i) identify and analyse design factors to be taken into account when building or purchasing self-reconfigurable robots aligned with green manufacturing requirements and (ii) reflect on domain experts' opinions about how such factors can be applied during the design process. To date, little, if any research has explored and matched design factors of self-reconfigurable robots with green requirements and thus this portion of the literature is weak with respect to theoretical foundations. The results of this research benefit (i) researchers looking for understanding and unpacking factors in green-aware self-reconfigurable robots to derive new hypotheses to be tested and to identify the areas of future research and (ii) practitioners looking for detailed advice on key design factors in the implementation of self-reconfigurable robots. The context in which this research is applied is the agriculture sector. We found that, in contrast to other fields such as industrial processes, the agriculture field has complex tasks which cannot be simply organised and modularised to be performed in simple actions via self-reconfigurable robots. Hence, we choose the greenhouse farming as it needs more investigation on how self-reconfigurable robots can be applied Roldán et al. (2018).

The rest of this paper is organized as follows. Section 2 provides some backgrounds on green manufacturing and self-reconfigurable robots. Section 3 presents the development of research hypotheses followed by the discussion of the research methodology. The paper ends with the discussion on future research, limitations, and a conclusion.

# 2 Background and literature review

The word *green* is used to reflect higher environmental awareness. In the context of manufacturing, it refers to adopting strategies and technologies that are more eco-efficient Deif (2011). The green manufacturing process covers the whole product lifecycle such as creating products/systems that consume less material and energy, substituting input materials (e.g. non-toxic for toxic, renewable for non-renewable), reducing unwanted outputs, and converting outputs to inputs (recycling).

It is commonly accepted that robots, among others, are a key integral part of manufacturing product lines to perform namely dumb, dangerous, dull, and/or dirty operations. A self-reconfigurable robot, a sub-class of robot systems, is one that is capable of adapting its functions and behavior to changing environments with minimum or without any external help Murata and Kurokawa (2007). Potentially, they are more adaptive than conventional robots to perform fixed tasks. A self-reconfigurable robot refers to the ability of the robot to adapt its functions at run-time based on changing user needs or operational environment, occurring intrusions or faults, and resource variability to cope with the complexity of today's environments. Self-reconfigurable robots can dynamically reconfigure, optimize, protect, and recover their components and functionalities while covering their internal complexity from users. Self-reconfigurable robots have been viewed as an effective way to move from a less green into a greener and more efficient manufacturing Murata and Kurokawa (2007). In general, there are a few motivations in designing self-reconfigurable robots Yim et al. (2007): (i) *Versatility*: Self-reconfigurable robots are able to disassemble and reassemble themselves to form new morphologies that are better suited for new tasks, for example changing from a legged robot to a snake robot and then to a rolling robot, (ii) *Robustness*: Self-reconfigurable robots can replace faulty components autonomously, leading to self-repair, (iii) *Low cost*: Self-reconfigurable robots can potentially lead to lower overall cost through creating and reusing copies of its components to brings economies of scale and mass production.

Reconfigurable robots have been identified as a key technology to keep the manufacturing industry green Pellicciari et al. (2015). Due to the fact that many of today's manufacturers are relying on robots in automating their product lines, improving green awareness of robots will subsequently lead to better green awareness of manufacturing Brossog et al. (2015). Hence this stream has received a significant attention from both researchers and industries seeking robots' design improvement. The development of green-aware self-reconfigurable robots is a challenging task Nielsen et al. (2017). That means, unleashing the capabilities of self-reconfigurable robots to achieve green-aware manufacturing implies a process including several stages, implementation activities, and principles for design and development of robots or reengineering existing ones. The development of robots should be analysed so as to select the adequate implementation techniques and technologies or a combination of them to achieve the desired green-awareness. This faces manufacturers with a challenge to design greener self-reconfigurable robotic systems. These challenges are better understood when a self-reconfigurable robot is characterized by its constituents.

There is a bunch of research on drivers of using green practices Vachon and Klassen (2006), models towards enabling green manufacturing Deif (2011), and frameworks/techniques to develop self-reconfigurable robots Edwards et al. (2009). Some other research intends to develop technical frameworks to increase the level of process automation Nakabo et al. (2009; Stasse et al. (2009), reduce the cost of production Quigley et al. (2011), and energy consumption Brossog et al. (2015); Brossog et al. (2014); Riazi et al. (2016). Such studies relating to this research have generally shown some insights of the effectiveness of robotic systems in the manufacturing process and propose useful techniques to reduce the energy consumption of robots associated with manual business processes.

Nevertheless, specific research on the quality design of self-reconfigurable robots to be accommodated in manufacturing to make them more green aware is more limited and inconclusive. Existing studies are rather technical-centric and focused on very specific robot design. As such it can hardly be generalized to get a theoretical ground. In designing and adopting self-reconfigurable robotic systems, incorporating the design factors influencing the environment is important Ahmad and Babar (2016); Pellicciari et al. (2015). From the perspective of empirical evidence, unpacking the influential factors in design and development of self-reconfigurable robots which subsequently positively/negatively contribute to improving green-awareness of manufacturing is hardly to find in the existing studies. The mixed results from the literature point to possible contingency factors that might be used in theory development in the effective green-aware robot design.

## 3 Hypothesis development

Like typical robots, a self-reconfigurable robot, regardless of its application domain, is a combination of two distinct components, i.e. hardware/physical for assembling and software for operations in a seamless way Jackson (2007). Software components aim to coordinate hardware components and to provide some advanced functionalities. Software components are typically embedded, real-time, distributed, and data-intensive and must meet specific requirements, such as safety, reliability, and fault-tolerance. Typically, developers implement software components through a software engineering lifecycle including common stages such as modelling, architecture designing and programming, testing and deploying. In developing green-aware self-reconfigurable robots, the aspects of software and hardware design are two key issues. In fact, designing green-aware self-reconfigurable robots can be viewed as balancing among software/hardware components design and their consequences on the environment. These two distinct aspects can be contradictory. For instance, the capability of a robot to reconfigure its components at the run-time may contrast high learning algorithm costs with low hardware setup costs or vice versa. We do not claim these aspects are complete, however, at this stage of our research, we found them as overarching for other independent fine granular variables in the literature. In the following, these constructs in relation to green manufacturing are delineated.

**Software design-time aspect.** The design-time aspect deals with writing codes and interfaces for robots in an efficient way. The research by Mahmoud and Ahmad (2013) argues that the way of coding for software systems influences the level of carbon emissions and hence has an indirect impact on the environment. In this manner, for example, Appasami and Suresh Joseph (2011) and Lo and Qian (2010) develop algorithms for compacting codes and data structures, reducing parallelism overhead, and routing data, which helps to control the power consumption by software applications. In the context of writing codes for robots, there are several design time aspects for robot software that should be taken into the account such as the choice of selecting algorithms for managing and allocating resources e.g. sensors, actuators, memory, processors, and robot motions. The generalised positive contributions of applying green practices during the architecture design have been extensively validated in the software engineering and accordingly can be extended to self-reconfigurable robots to postulate the positive impact of applying green practices. If applying the design-time principles for the robot's software is viewed as being relatively important to optimise energy usage and minimise negative impacts on the environment, they are required to be adequately addressed in designing the software components of a robot. Hence, we derive the following hypothesis:

*H1a: Applying green practices during the development of a self-reconfigurable robot's software components positively contributes to greener manufacturing.*

**Software run-time design aspect.** This aspect deals with implementing capabilities in software components of robots to cope with uncertain operating conditions of the environment. Software components which perform robots' functions may need to change their behaviour due to occurrence of faults, resource limitation, or user preference. Self-reconfigurable robots can dynamically reconfigure, optimize, protect, and recover their components and functionalities. Furthermore, the run-time aspects of a robot's software deal with enabling software components to monitor their resource utilisations at the run-time and - if necessary - to improve it. For example, a robot may need to change a resource intensive image processing algorithm in its sensor to a low-energy consuming algorithm if the robot battery is getting low. The run-time adaptation in self-reconfigurable robots can be in the form of changes in the physical environment (e.g. the loss of a power source), changes in the underlying hardware and network infrastructure (e.g. such as a device failure), changes in the available software resources (e.g. update to a behaviour control component) and changes in the scenario goals (e.g. a switch from object following to wall mapping) Edwards et al. (2009). For example, cloud computing-based self-reconfigurable robotics can discover and utilise distributed, virtually unlimited, and powerful hardware/software components on the internet Diaconescu and Wagner (2014; Riazuelo et al. (2014). This means, a robot's components can be dynamically swapped or combined with new ones that have lower energy consumption rates. This is realised through the learning ability of components (e.g. energy-efficient motion planning using machine learning techniques) to acquire knowledge from the environment and to learn how to adapt themselves to new situations. Nevertheless, the fundamental trade-off lies between rearranging the connectivity of a robot's components in order to adapt to new situations and the energy consumed for the execution of tasks by the robot Hu et al. (2012). If dynamic component replacement and resource allocation are not provisioned quickly, effective energy consumption may not be achieved. Moreover, the robot capability in learning and adaptation is constrained by its processing power, storage space, and the number and type of sensors it carries. This may cause developers to ignore incorporating green practices into the run-time aspects of a robot design. The usefulness of applying run-time aspects of the robot's design aspect may have a positive effect on energy consumption. Hence, we hypothesize:

*H1b: The capability of a robot in changing its software components at run-time for efficient energy consumption can positively contribute to greener manufacturing.*

**Hardware design aspect.** Beyond the robot software design, the hardware design may subsequently incur carbon emissions during operation and wasting. For example, a robot may rely on traditional energy sources such as electricity or petroleum oil which may not be an effective energy source in all conditions and presents threats of pollution to the environment Grau Saldes et al. (2016). In addition, one should not be surprised that the raw physical material used for building hardware components of robots is also a major factor for carbon emissions, recycling and even the safe disposal phase of robotic components after their retirement. Hardware components such as processors, memories, actuators, sensors, camera, and other mechanical components, that are controlled and manipulated via software components, can be procured from different suppliers. The substances used in these components may cause excessive carbon emission during high workload Kannan et al. (2014). Hence, eco-design principles should be integrated with the robot hardware design. For example, in the selection of robot hardware components it is advised to find suppliers avoiding the use of illegal materials, complies with the environmental laws, and offering components that do not contain dangerous and toxic substances. Following this logic, we define that:

> H2a: *The use of environmentally friendly material for building hardware components of robots positively contributes to greener manufacturing.*

The research model presented in Figure 1 is preliminary and parsimonious. It integrates predictors for designing green-aware self-reconfigurable robots from different design perspectives. At this stage of the research, we do not claim the presented research model is exhaustive and there may be other predictors of a typical robot design which are left unexplored for future research.

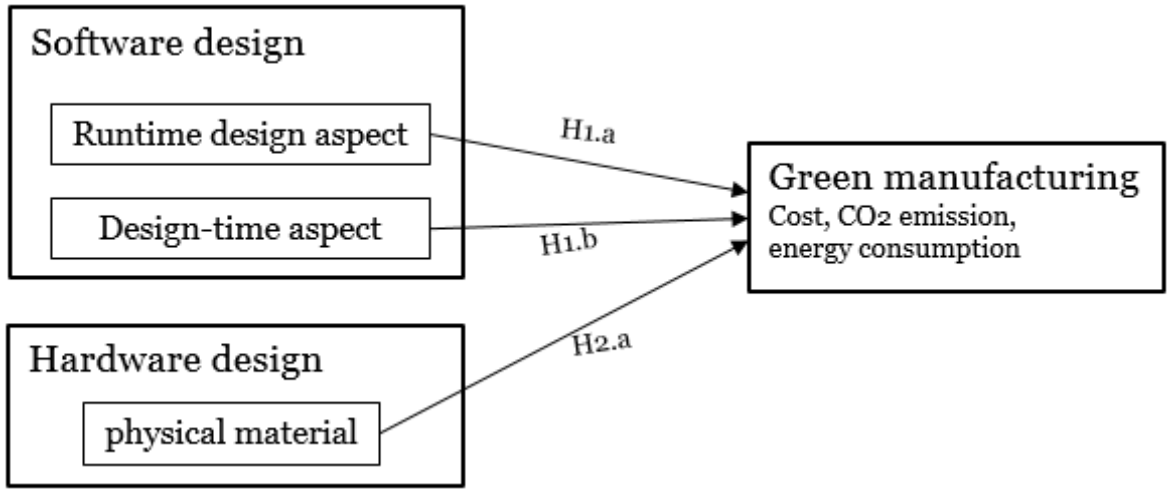

*Figure 1: Research model of green-aware robot design for manufacturing*

## 4 Research methodology

The next step involves testing the presented research model. We plan to validate the research model in a two-phase study. Firstly, we will conduct semi-structured interviews assessing practitioners' views of the importance of the constructs in addressing green manufacturing requirements. This enables us to initially investigate our contingency constructs for the external validity. Then, our hypothesized model of self-reconfigurable robot design will be tested empirically using a longitudinal survey of engineers with experience in building self-reconfigurable robots for the agriculture sector. Using self-reconfigurable robots in agriculture is aimed at building intelligently efficient farming systems performing repetitive and cumbersome operations such as seeding, weeding, irrigation, culling fruits, and removing those failed to thrives Roldán et al. (2018).

**Survey design and construct measurement.** We have already designed a web-based survey using Qualtrics (Snow and Mann, 2013), a survey designer software able to capture survey results, which is ready to be sent to participants. Our survey has two parts. In the first part, the common terminologies used in the survey including green manufacturing and self-reconfigurable robots are introduced. The second part constitutes a list of questions asking respondents to rate the level of the importance of design aspects in achieving green-aware self-reconfigurable robots. For the rating of importance level of each construct, we use a 1–7 Likert scale: 1 for not-at-all-important, 2 for very unimportant, 3 for somewhat unimportant, 4 for neither important nor unimportant, 5 somewhat important, 6 important, and 7 extremely important. The constructs defined in the research model are measured using multiple-items and seven-point scales. For each construct, a definition including an example is provided to help respondents to get a better understanding of the aspect and the scope of the questions. The respondents can provide additional comments and suggest any missing green design that had to be covered in the model. As mentioned in Section 3, the software design is a two-dimensional construct comprising of two sub-dimensions, i.e. design-time and run-time aspects. The software design-time construct of the self-reconfigurable robot is measured using Mahmoud and Ahmad (2013), Appasami and Suresh Joseph (2011), and Lo and Qian (2010). The software run-time construct is measured using Diaconescu and Wagner (2014; Riazuelo et al. (2014) and Hu et al. (2012). The hardware design construct is measured using variables adapted from Kannan et al. (2014).

**Data collection and analysis**. This voluntary survey will be public and shared through sending email to qualified experts who have a profile in both green manufacturing enablement and robotic systems. An eligible respondent to participate in the survey is a person who has real-world experience in enabling green awareness using robotic systems. Prior to the recruitment of each participant and through an inquiry email, we ask each participant if he/she has had experience on operationalisation of green practice through utilising capabilities of self-reconfigurable robotic systems. Once the response and profile are found satisfactory, an invitation letter along with the link to the survey is sent through a second email. We also use the purposeful sample technique Kitchenham and Pfleeger (2002) to increase the credibility of the collected responses. That is the first respondent will help us to identify other respondents who may have a satisfactory profile to complete the survey. To test the validity of the model measurements confirmatory factor analysis will be used.

# 5 Research summary, contributions, and limitations

A green aware self-reconfigurable robot design can minimise the overall environmental impact and mitigate the manufacturer's future environmental risks. Understanding the impact of design aspects of self-reconfigurable robots is of practical importance in achieving more environmentally friendly manufacturing and how it can be effectively addressed. We have proposed that self-reconfigurable robot design is involved with the three main constructs of design-time, run-time design, and hardware design. We discussed that all these constructs are equal, even though there can be some trade-offs, for example in terms of cost, or contradictory situations among the constructs. We developed propositions to be attested in further research. More research required to identify the contingencies where self-reconfigurable design can more effective.

This study theorises and empirically attest key factors in designing green-aware self-reconfigurable robots. Through addressing this research question we hope to make both theoretical and practical contributions. Given the rapid growth of utilizing robots in automating manufacturing systems to reduce operating costs and to increase the productivity, we expect the following contributions. First, our research aims to theoretically unpack the critical factors that are to be balanced and taken into account when designing self-reconfigurable robots for making advanced green-aware manufacturing systems. We conceptualise the green-aware design that accordingly has an effect on advanced green manufacturing and Industry 4.0. This connection between self-reconfigurable robots and manufacturing systems can demystify why some manufacturers are better in the realisation of green strategies compared to other others. Secondly, the results of this research provide insights into the important design factors for manufactures interested in utilising or improving existing robotic systems. Moreover, we strengthen the empirical evidence in the literature via collecting and analysing data from the industry sector. So far empirical findings regarding the impact of effective robot design in addressing green manufacturing requirements have not been received attention by researchers. This research also contributes to engineering practice to design or acquire green-aware robots. That is, we expect to demonstrate how companies are able to govern their data analytics related resources to become innovative using available data sources.

Our research findings should be seen in view of some limitations. Firstly, besides of identified constructs presented in the research, there might be many others and hence we do not claim the model completeness. Instead, we developed the first step towards better understanding of efficient self-reconfigurable robot design in the light of green manufacturing. Therewith, we inform scholars and practitioners aim at improving existing or building new green-aware robotic systems. The second limitation is that our data will mainly be collected from agriculture companies designing self-reconfigurable robots. Future research should consider data collection from other industry sectors.